\documentclass[10pt,twocolumn,letterpaper]{article}

\usepackage{cvpr}
\usepackage{times}
\usepackage{epsfig}
\usepackage{graphicx}
\usepackage{amsmath}
\usepackage{amssymb}
\usepackage{dirtytalk}

\cvprfinalcopy % *** Uncomment this line for the final submission

\def\cvprPaperID{****} % *** Enter the CVPR Paper ID here

\begin{document}

%%%%%%%%% TITLE

\title{Distill-2MD-MTL: Data Distillation based on Multi-Dataset Multi-Domain Multi-Task Frame Work to Solve Face Related Tasks}

\author{Sepidehsadat Hosseini\\
Seoul National University\\
Seoul, Korea\\
{\tt\small sepid@snu.ac.kr}
\and
Mohammad Amin Shabani\\
Seoul National University\\
Seoul, Korea\\
{\tt\small aminshabani@snu.ac.kr}
\and
Nam Ik Cho\\
Seoul National Univeristy\\
Seoul, Korea\\
{\tt\small nicho@snu.ac.kr}
}
\maketitle

% Enter the paper's authors in order
% \addauthor{Name}{email/homepage}{INSTITUTION_CODE}

% These are not defined in the style file, because they don't begin
% with \bmva, so they might conflict with the user's own macros.

%-------------------------------------------------------------------------
% Document starts here

\begin{abstract}

We propose a new semi-supervised learning method on face-related tasks based on Multi-Task Learning (MTL) and data distillation. The proposed method exploits multiple datasets with different labels for different-but-related tasks such as simultaneous age, gender, race, facial expression estimation. Specifically, when there are only a few well-labeled data for a specific task among the multiple related ones, we exploit the labels of other related tasks in different domains. Our approach is composed of (1) a new MTL method which can deal with weakly labeled datasets and perform several tasks simultaneously, and (2) an MTL-based data distillation framework which enables network generalization for the training and test data from different domains. Experiments show that the proposed multi-task system performs each task better than the baseline single task. It is also demonstrated that using different domain datasets along with the main dataset can enhance network generalization and overcome the domain differences between datasets. Also, comparing data distillation both on the baseline and MTL framework, the latter shows more accurate predictions on unlabeled data from different domains. Furthermore, by proposing a new learning-rate optimization method, our proposed network is able to dynamically tune its learning rate.

\end{abstract}

%-------------------------------------------------------------------------
\section{Introduction}
\label{sec:intro}

\label{sec:intro}

Deep learning based frameworks have shown outstanding performance in many computer vision problems. However, as the target task becomes multi-faceted and more complicated, sometimes unduly large dataset with stronger labels is required. Hence, the cost of generating desired labeled data for complicated learning tasks is often an obstacle, especially for multi-task learning. Therefore, studies on semi/self/omni-supervised learning are getting attention recently because they can obviate such strong labeling. In the most semi-supervised learning methods, they exploits part of annotated data and considers the rest as unlabeled~\cite{sheikhpour2017survey,classic_semi_1}. Recently a new regime of semi-supervised learning has been proposed called as omni-supervised learning~\cite{radosavovic2017dd}. In the omni-supervised learning, the learner uses as much labeled data as possible and also uses an unlimited amount of unannotated data from other sources. 

In this paper, we propose a data distillation framework on weakly labeled datasets to help to improve the multi-task learning on facial expression recognition. Previous works on distillation adopted omni-supervised learning methods~\cite{radosavovic2017dd} which used unlabeled auxiliary datasets. However, we argue that instead of feeding the network with unlabeled images for providing a new target labeled dataset, we can use datasets from other related tasks as weakly labeled images. By doing so, we can train the network in the manner of multi-task learning (MTL) and then use the trained network to produce the target labels for the related tasks' datasets. Then, similar to \cite{radosavovic2017dd, furlanello2018born}, we retrain the network in a single task manner with the union of the original and the newly labeled datasets. By doing so, we can benefit from making the network familiar with the features of the new datasets and having a more powerful teacher for data distillation.

The MTL based methods can be much helpful to achieve the real-time visual understanding of a dynamic scene, as they are able to perform several different perceptual tasks simultaneously and efficiently. In the exemplary MTL methods, we need to consider constructing a dataset that contains all the labels for different tasks together. Without such a dataset, training the multi-task network in a common approach will result in a negative effect due to the cross-dataset distribution shift. To the best of our knowledge, the first work which mentioned this problem is StarGAN~\cite{choi2017stargan} proposed by Choi \etal.  Their model can simultaneously be trained on different datasets by alternating between different datasets. However, the alternating scheme still has the cross-dataset distribution shift problem, and the network cannot be applied to datasets with different domains. Recently, Guosheng Hu \etal~\cite{eccvhu} addressed this issue by proposing the trace norm-based knowledge sharing. In their method, multiple networks, one for each task, are stacked horizontally together to form a one-order higher tensor. Then, by using a tensor trace norm regularizer, they share knowledge between these networks. In comparison with \cite{eccvhu}, our method is simpler, easier to implement, and more efficient in both aspects of memory and computation.
.

 \begin{figure*}[t]

      \begin{center}
          \scalebox{0.15}{\includegraphics[scale=0.375]{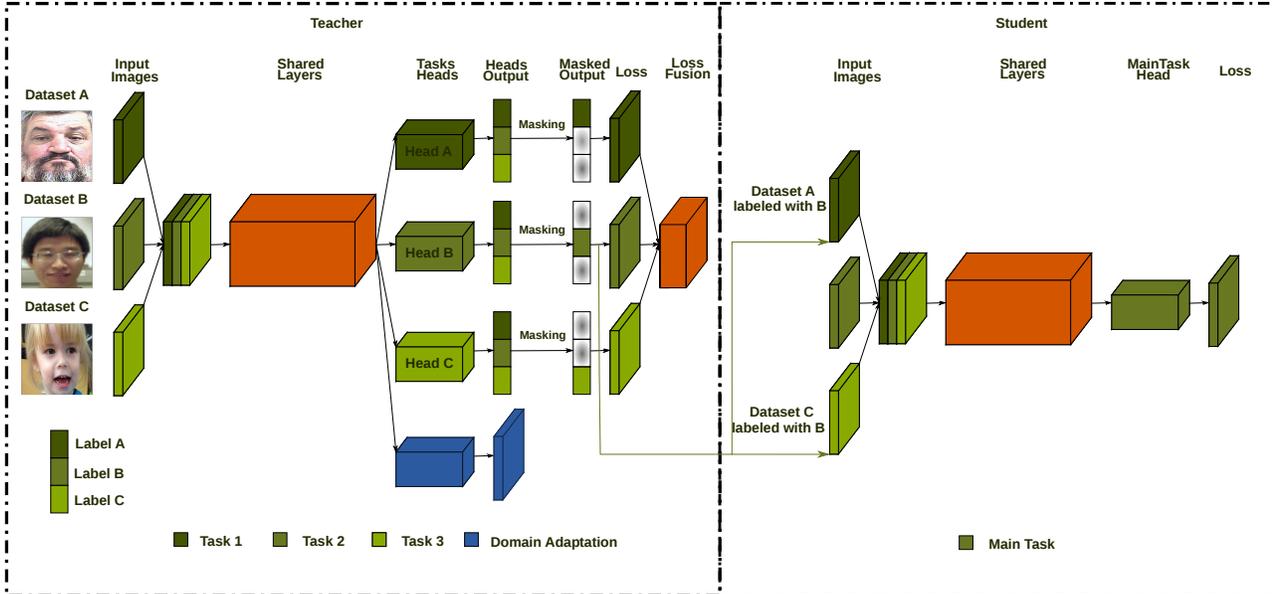}}
      \end{center}
      \caption{The proposed method. Right: the first step of training using the proposed 2MD-MTL network (teacher). Left: the second step of training using a simple single task network 
      (student) with labels produced by the 2MD-MTL network (teacher). }
      \label{fig::batch}
  \end{figure*}

In our multi-dataset multi-task method, we use a single network instead of multiple networks and employ a mask vector for the loss functions. Therefore, we may focus on the input module by creating a combination of the data from different tasks in each input batch of the training procedure. By doing so, we use just a single multi-task network, similar to the common models. Also, for each input batch, we evaluate the loss function related to the task of each image as shown in Figure~\ref{fig::batch}. It makes all of the tasks be involved in each backpropagation of the network which significantly helps to reduce domain shift problem. To decrease the domain shift problem more, we add metric learning based domain adaptation to our model. Furthermore, we propose a new method to adjust the learning rate schedule dynamically using loss reduction rate as feedback.

To evaluate our proposed framework we divide our experiments into two major parts, in all these steps we use Facial Expression Recognition as main task and two more simpler tasks, age, and gender estimation, as auxiliary tasks. In our first part, we show how our framework can improve network performance both on main and auxiliary tasks. On the second part, we conduct our experiments on to measure our network generalization, therefore in this part we train our network on a certain domain and then test it in different domain.

\section{Related Work}
%-------------------------------------------------------------------------
\noindent\textbf{Semi-Supervised Learning.} 
Zhu\etal ~\cite{zhu2006semi} and Sheikhpour \etal~\cite{sheikhpour2017survey} have done comprehensive surveys on semi-supervised learning methods.  The first trial on self semi-supervised learning was based on the soft self-training technique~\cite{classic_semi_1},  which is to predict labels of unannotated data. Then those labels are used to train itself, which is known as one of the simplest and commonly used approaches in semi-supervised learning. Recently, many approaches attempt designing deep learning based semi-supervised frameworks~\cite{DBLP:booktitles/corr/RasmusVHBR15, LaineA16, radosavovic2017dd}. Rasmus \etal~\cite{DBLP:booktitles/corr/RasmusVHBR15} proposed the Ladder network-based method, by exploiting unsupervised auxiliary tasks. Laine \etal \cite{LaineA16} annotate unlabeled data using the outputs of the network-in-training under different conditions such as regularization input augmentation. In Omni-supervised method~\cite{radosavovic2017dd}, they use knowledge distillation from larger data, in the other word their model generates annotations on unlabeled data using a model trained on large amounts of labeled data. Then, they retrain the model using the extra generated annotations.
\vspace{0.3cm}

\noindent\textbf{Facial Expression Recognition.}  Facial Expression Recognition (FER) has also attained increasing attention recently. 
Li and Deng~\cite{Li2018DeepFE} published a survey on the deep facial expression recognition methods. Recently, Yang \etal~\cite{yang2018facial} proposed to recognize facial expressions by extracting information of the expressive component through a de-expression learning procedure, called De-expression Residue Learning (DeRL). Zhang \etal~\cite{zhang2018joint} proposed joint pose and expression modeling by disentangling the expression and pose from the facial images and produce images with arbitrary expressions and poses using a new discriminator and a content-similarity loss for generative adversarial networks. Zeng \etal~\cite{Zeng_2018_ECCV} addressed the inconsistency between FER datasets for the first time by proposing an Inconsistent Pseudo Annotations to Latent Truth (IPA2LT) framework to train a FER model from multiple inconsistently labeled datasets and large-scale unlabeled data. Our method can be considered a generalization of this work because we can use datasets with inconsistent labels instead of datasets with different-task labels. 

\noindent\textbf{Multi-Task Learning.} Multi-task learning has demonstrated performance improvement in several computer vision applications such as facial landmark detection~\cite{multi_landmark} and human pose estimation~\cite{multi_pose}. 
The primary intuition behind Multi-Task Learning (MTL) is how humans apply their knowledge and skills obtained from other tasks on more complicated tasks. There are different methods to exploit MTL: joint learning, parallel multi-task learning with auxiliary tasks, and continual learning are a few examples of MTL based methods. 
The parallel multi-task based methods integrated different tasks contemporaneously, which has been widely deployed in face-related tasks~\cite{eccvhu,multi_face_rep}.
\vspace{0.3cm}

\noindent\textbf{Knowledge and data distillation.} There are a large number of researches attempt to transfer knowledge from a teacher model to a student model. Romero \etal~\cite{DBLP:journals/corr/RomeroBKCGB14} proposed FitNets, a two-stage strategy to train networks by providing \textit{hint} from the teacher middle layers. Knowledge Distillation (KD) proposed by Hinton \etal~\cite{44873} leverage the predictions of a larger model as the \textit{soft target} to better training of a smaller model. After that, Chen \etal~\cite{NIPS2017_6676} improved the efficiency and the accuracy of an object detector by transferring the knowledge from a powerful teacher in case of model architecture or the input data resolution to a weaker student. Zagoruyko \etal~\cite{Zagoruyko2017AT} proposed several ways to transfer the attention from a teacher network to a student. Polino \etal~\cite{polino2018model} proposed quantized distillation to compress a network in terms of depth by using knowledge distillation. Furlanello \etal~\cite{furlanello2018born} used knowledge distillation on a student the same as the teacher to improve the performance of the networks by teaching selves.

 Inspired by knowledge distillation, Radosavovic \etal~\cite{radosavovic2017dd} proposed data distillation to tackle omni-supervised learning. They generate annotations for unlabeled data by using a trained model on a labeled dataset and then retrain the model on the union of these two datasets to improve the accuracy. There are also other works trying to use unlabeled data to retrain the model \cite{yarowsky1995unsupervised,4270073, Chen-2013-7815, LaineA16}. Gupta \etal \cite{GuptaHM15} proposed a method to transfer supervision between different modalities which needs unlabeled paired images.  Laine and Aila~\cite{LaineA16} proposed to use ensemble from different checkpoints with different regularizations and input augmentations.

\noindent\textbf{Domain Adaptation.} Saenko \etal~\cite{DA1} was one of the first researchers who proposed a method to solve the domain shift problem. More recent works are based on deep neural network aiming to align features by minimizing domain gaps using some distance function~\cite{domain2,domain3}. In these methods, domain discriminator trains to distinguish different domains while the generator tries to fool discriminator through the learning of more general representation and features.

\section{Proposed Method}

\subsection{MTL learning}
Suppose we have $t$ tasks $T_i$ for $1\leq i \leq t$, and $d$ datasets $D_j$ for $1\leq j \leq d$ in which each dataset contains labels for a subset of the $t$ tasks. Without loss of generality, suppose the target task is $T_1$ and at least one of the datasets contains the related labels for the target task. By defining a multi-task network $N_m$, we train $N_m$ with the datasets $D_j$ in a multi-dataset multi-domain multi-task (2MD-MTL) manner. 
    
To be more clear, instead of training the network with alternating inputs from each dataset, we construct an input batch of size $b = t\times \hat{b}$ as a combination of $\hat{b}$ images from each task $T_i$. Therefore, by evaluation of the network $N_m$ on the input batch, we will have a matrix $L$ of size $b\times t$ related to the loss functions of the different tasks, in which cell $l_{i,j}$ means the loss value for the $i$-th image and the $j$-th task. Now, we construct a mask matrix $M$ of the same size by putting $m_{i,j} = \alpha_j$, where $\alpha_j$ is equal to the coefficient of the loss due to the task $T_j$, if the $i$-th image contains the label for task $T_j$ and 0 otherwise. Then, the final loss will be equal to the dot product of these two matrices. In other words, we use all the tasks parallelly in the network by considering only the valid loss values at the end.

The loss function in multi-task learning is generally defined as $L=\Sigma_i\omega_iL_i$, where $L_i$ is a Loss function and $\omega_i$ is a scalar coefficient for the $i$-th task respectively. In most of the cases, it is challenging to find the best value for each $\omega_i$ which not only need huge efforts and extensive experiments but also decrease network generalization. We use the gradient normalization~\cite{gradnorm} to solve the loss balancing problem, which obviates the expensive time-consuming grid search for tuning the $\omega_i$s.

Figure~\ref{fig::batch} shows the proposed framework, where we use VGG-16~\cite{vgg16} network as the baseline, all tasks are sharing convolutional layers (5 convolutional blocks), and each task has its own Fully connected layers and also their own loss as their head.

 %%%\begin{figure}[t]
%
 %     \begin{center}
  %        \scalebox{0.4}{\includegraphics[scale=0.27]{datasetss.eps}}
   %   \end{center}
    %  \caption{Diverse-domains - diverse-tasks datasets}
     % \label{fig:datasets}
  %\end{figure}     %-------------------------------------------------------------------------
    
\begin{figure}[t]

\centering
    \scalebox{0.38}{\includegraphics[scale=0.27]{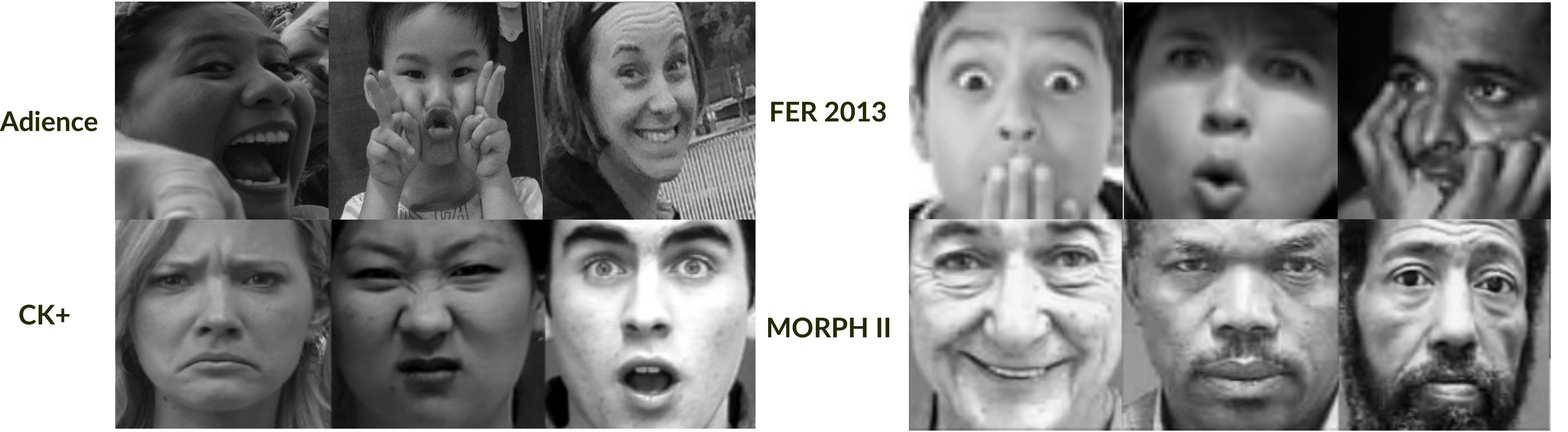}}
    
 \caption{Diverse-domains - diverse-tasks datasets} 
 \label{fig:datasets}

\end{figure}

While the features learned above on multiple tasks will be more general-purpose ones than those learned on a single task, there may be still a problem if the dataset domains are so different. For example, Figure~\ref{fig:datasets} shows some images used in our experiments, where the images from MORPHII and Casia datasets are frontal images while the photos from the Adience dataset are mainly wild. On the other hand, MORPHII subjects are photos of prisoners' who don't show that many emotions while Casia is emotion dataset. In order to minimize the domain gap between different datasets and also extract more generalized features, we use metric learning based discriminator. 
    
\begin{figure}
\centering
  \scalebox{0.42}{\includegraphics[scale=0.375]{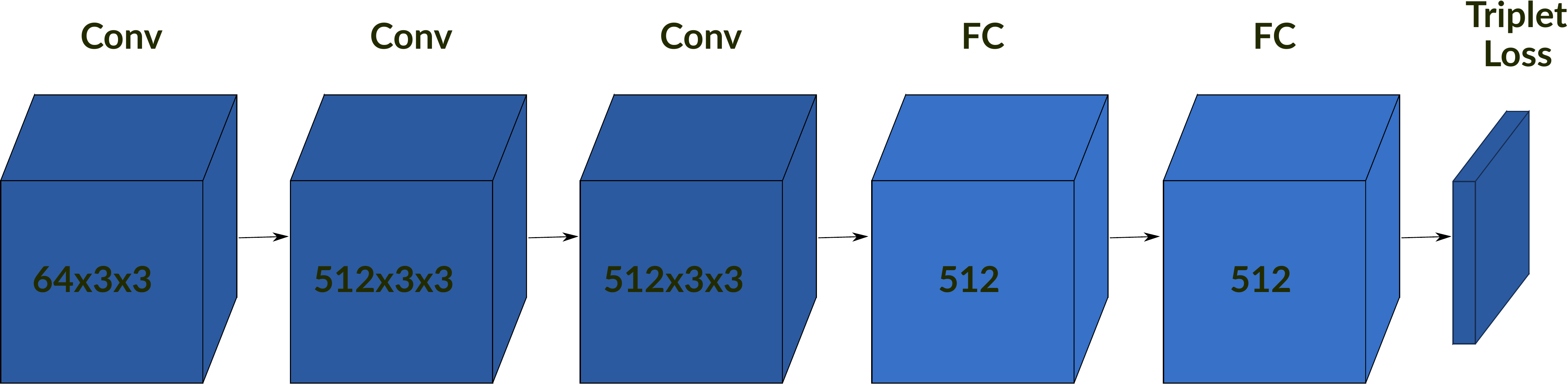}}

 \caption{ Discriminator Head.\newline}
  \vspace{0.3cm}
 \label{fig:dis_head}

\end{figure}

In our proposed method, we add the discriminator head shown in Figure~\ref{fig:dis_head} after the shared layers. Then, we apply a triplet loss which aims to pull samples belonging to the same dataset into nearby points on a manifold surface and push samples from different datasets
apart from each other. The labels of the training images for the discriminator's head can be easily provided by the dataset to which they belong (for example 0: age, 1: gender, and 2: emotion). Then, they are selected and formed into triplets as $T_i=(x^a, y^p, y^n)$, where $x^a$ and $y^p$ are the anchor and positive samples respectively which belong to one dataset and $y^n$ is the negative sample which belongs to another.
Then, we train the discriminator head to decrease the triplet loss (Eq.~\ref{eq:1}) and the rest of the network to minimize the total loss (Eq.~\ref{eq:2}) where $N$ is the number of tasks and $L_i$ is the loss function per task. In the other word, we train network in the way that the discriminator's head is not able to distinguish between datasets, while still the other heads being able to extract informative feature for all the task, therefore shared layer features will be generalized on all tasks and domains.

\begin{equation}
L_{DA}=\Sigma^T_{t=1}[||x^a_t -y^b_t||^2_2 - ||x^a_t -y^b_t||^2_2 +0.2]
\label{eq:1}
\end{equation}

\begin{equation}
L_{Total}=\Sigma^N_{i=1} \omega_i L_i - L_{DA}
\label{eq:2}
\end{equation}

Using the triplet loss for the discriminator's head can help us to overcome the class imbalance problem due to the different sizes of datasets. For example, the number of MORPHII images is one order of magnitude greater than the images in CK+. Therefore, without considering a solution for the class imbalance, the discriminator will be biased to MORPHII based on a large number of images in that class.


\subsection{Data Distillation}

Hinton \etal~\cite{44873} proposed knowledge distillation (KD) in order to transfer the knowledge from a cumbersome teacher model to a smaller student model. They use the class probabilities predicted by the teacher model as a \textit{soft target} to guide the student model. Furlanello \etal in born-again neural networks~\cite{furlanello2018born} show that we can also adopt student network architecture as the teacher in order to improve the model by guiding itself. Radosavovic \etal~\cite{radosavovic2017dd} apply this idea to omni-supervised learning. They showed that by using a trained model with a labeled dataset, we can generate labels for an unlabeled dataset by applying the model on multiple transformations of the input images and aggregate the results as the \textit{hard labels} similar to the ground truth labels. It has been shown that the aggregation will improve the results in~\cite{krizhevsky2012imagenet,szegedy2014scalable,felzenszwalb2010object}. Comparing to the previous methods, we believe that using weakly labeled datasets in a multi-task learning manner instead of an unlabeled one has advantages especially when the distributions of the labeled datasets and the unlabeled one is highly different. 
    
For example, in the case of face-related tasks,  if we have a dataset consists of images in domain \say{A} in a specific task \say{X} (such as facial expression recognition), and  we have a datasets of images in domain \say{B} which they have different features with images in domain \say{A} and they are labeled by the other task \say{Y} ( such as age estimation). Then if we want to use a model trained on domain \say{A} to estimate the facial expression of domain \say{B}, the model which is trained only on a specific domain probably will suffer from the differences of features between the domain and won't show a good performance. Therefore, the proposed method in \cite{radosavovic2017dd} cannot produce good labels without adopting the new domain.

Therefore, in our method, we used our proposed trained MTL framework, which can learn more general features, to generate Unknown labels for all datasets (Figure~\ref{fig::batch}). For examples, if dataset \say{A} has been annotated for task \say{X} but not task \say{Y} and \say{Z} we use our MTL network to generate \say{Y \& Z} labels for task \say{A}, then by doing so, we can generate more accurate predictions and can train our network in the classic MTL manner. 

\section{Datasets}
We consider 5 facial related datasets to evaluate our method. 

\noindent{\bf CK+~\cite{ck+}} is one of the constrained datasets widely used for FER. It contains 593 video sequences from 123 persons. The sequences start from neutral faces and shift to one of anger, contempt,
disgust, fear, happiness, sadness, and surprise expressions peak. Among these 593 sequences, only 327 sequences from 118 persons are labeled to those seven expressions.
    
\noindent{\bf Oulu Casia~\cite{casia}} contains 2,880 sequences of 180 subjects, in six different expressions (anger, disgust, fear, happiness, sadness, and surprise) per subject. Similar to CK+ each sequence starts from a neutral face and gradually shows the expression. Following other researches, we also use only images under visible light and strong illumination condition.
    
     \noindent{\bf FER2013~\cite{FER}} is annotated with seven basic facial expressions (0=Angry, 1=Disgust, 2=Fear, 3=Happy, 4=Sad, 5=Surprise, and 6=Neutral), which contains about 32K images, 28.5K for training and 3.5K for the test. All pictures in this dataset are collected automatically by the Google image search API which is one of the frequently used unconstrained datasets.
    
    \noindent{\bf MORPHII~\cite{Morph}} is one of the most popular large-scale age estimation datasets created by the Face Aging Group at the University of North Carolina. It contains 55,134 images of 13,000 subjects with about three images per subject, age ranging from 16 to 77 year. The images in this dataset are mainly frontal.
    
    \noindent{\bf Adience ~\cite{adience}} compared to MORPHII which contains frontal and constrained images, has been captured from Flicker.com albums. Hence, they are totally unconstrained and no manual filtering has been applied, which makes them a good representation of the real world. It consists of 26K facial images of 2,284 identities.  
    
\section{Experiment and Results}
We conduct experiments mainly on Facial Expression recognition, and we divide our analysis into two main parts. In the first part, we compare our network with a single-task baseline when both training data and test data are from the same domain and the same dataset (Sec.~\ref{Semi-Supervised MTL}). In Sec.~\ref{Semi-Supervised MTL}, we also evaluate our network on the auxiliary tasks (age and Gender estimation) to prove that not only our network shows the better result on the main task (FER), but also simultaneously improves the performance of those auxiliary ones. Following the previous works, we use 10-fold cross validation protocol for all the experiments on both CK+ and Casia datasets. We repeat each experiment 10 times and report the average result. At the last part, we compare our result when test data are from a different dataset from other domain in order to evaluate the generalization of our proposed network (Sec.~\ref{Semi-Supervised cross domain}). In this part we evaluate our network on FER2013  which has been already divided to train part and private test set by publishers~\cite{FER}, we follow their protocol and evaluate our network on private test part of FER2013.

 In the experiments, all the images are resized to $48\times48$, and batch size to 128 which is divided into three parts 32, 32, and 64 for age, gender, and emotion datasets respectively, and we train our network for 200 epochs per each experiment. We utilize conventional data augmentation in the form of random sampling and horizontal flipping. To adapt VGG-16 network to our $48\times48$ input, we omit the last pooling layer right after VGG-16 5th block.

For optimization, we used Momentum optimizer and fix the momentum to be 0.9. We use two different methods in order to adjust the learning rate, the first one is the classic method where the learning rate starts from $10e-2$ and dropped exponentially, in the second method as the recent researches demonstrated that instead of monotonically decreasing the learning rate, vary learning rate cyclically will cause improve in performance without a need to tune and often in fewer iterations~\cite{cylce_lr}, we proposed a dynamic learning rate, in our method network gets feedback from loss difference between iteration and if it seems not decreasing enough it will decrease learning rate ($lr_{T+1}=lr_{T}*0.1^{(current-training- step/10^k)}$) and in each $k_th$ times (lets call it \say{cycle}) that this situation happens instead of decreasing learning rate, we will increase it to the initial learning rate call it as $lr_{MUX}$, as a result, the minimum learning rates in each cycle, step $k-1$ in each cycle, will be equal to the same value as if the learning rate has been exponentially decreased, in our experiments we set \say k to 5.

\subsection{Semi-Supervised MTL}
\label{Semi-Supervised MTL}
\begin{figure}
\centering
  \scalebox{0.5}{\includegraphics[scale=0.6]{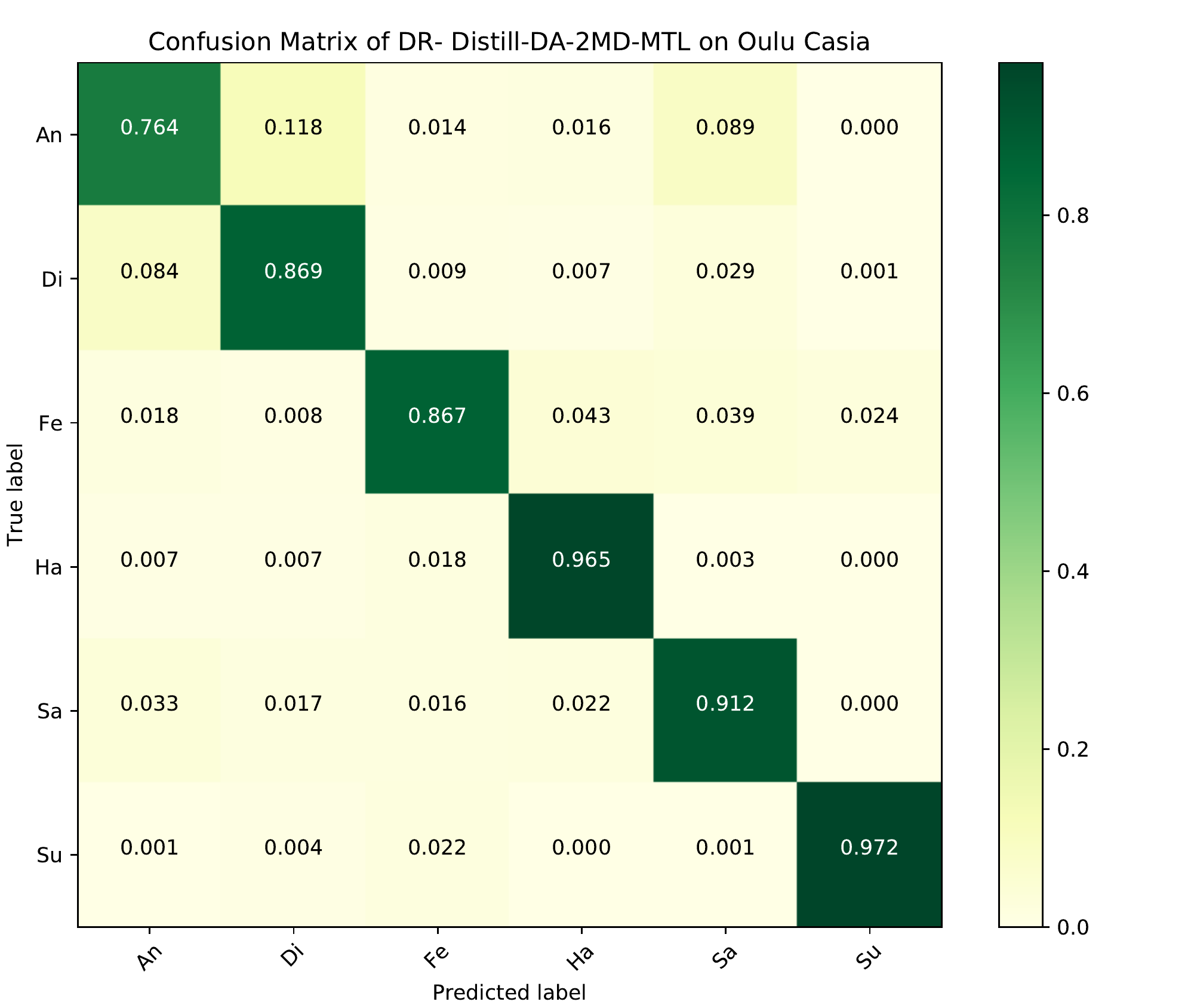}}

 \caption{Confusion Matrix from DR- Distill-DA-2MD-MTL on Casia). The darker the color, the higher the accuracy.}

 \label{conf}
\end{figure}

In this section we use three tasks; age estimation on MORPHII, gender estimation on Audience and FER on CK+, Oulu Casia. For facilitating the age estimation task, we divide it into two classes, those who are younger than 38 years old and those who are older than 43 years old and we ignore the rest. Then we evaluate our network on CK+ and Oulu Casia respectively. To have a fair compression with state of the arts as they mostly pre-trained their network~\cite{FN2EN,Li2018DeepFE,LI2017135,eccvhu,yang2018facial}, we also follow their method and pre-trained our network on LSEMSW same as~\cite{eccvhu} and then fine tune our network on CK+ and Casia  While pre-training we didn't change the age and gender datasets. The result has been shown in Table~\ref{table:1}, Confusion matrices are also has been shown in Figure~\ref{conf}. Moreover, we evaluate our network on MORPHII (age Estimation) and Adience (gender Estimation), while we use all images of Oulu Casia for training the FER. We divide both age and gender to two parts of train and test with a portion of 4 to 1.\newline The result has been shown in Tabel~\ref{table:3}. ``DA'' prefix indicates networks with Domain Adaptation, ``Distill'' for the network using knowledge distillation and ``DR'' for the network being trained using proposed dynamic learning rate.  As the results show, the proposed method not only gets a great improvement over the baseline by exploiting the information of the other datasets from other tasks but also it works better than other multi-task approaches ~\cite{eccvhu,choi2017stargan} and other states of the art techniques.

\begin{table}[h]
\begin{center}
\scalebox{0.8}{
\begin{tabular}{|l | {c} |c| }
\hline
 Method & FER on Oulu-Casia & FER on CK+\\\hline  
HOG 3D~\cite{HOG3D} &70.63 &91.44 \\

IPA2LT~\cite{Zeng_2018_ECCV} & 61.02 & 91.67  \\
FN2EN ~\cite{yang2018facial} &87.71 &96.8 \\
DeRL~\cite{yang2018facial}&88.0 &97.30 (7classes) \\

\hline
RN+LAF+ADA ~\cite{eccvhu}& 87.1 &  96.4\\
Star-Gan~\cite{choi2017stargan} & 84.3 & 91.6\\
\hline
Baseline & 83.6 &  89.4 \\
2MD-MTL  & 86.83 & 93.51  \\
DA-2MD-MTL & 87.1   &  93.4  \\
Distill-Baseline& 84.1 & 89.2 \\
Distill-2MD-MTL&  89.13 & 94.5    \\
Distill-DA-2MD-MTL&  89.3 &   96.73 \\
DR-Distill-DA-2MD-MTL& \textbf{90.1}  & \textbf{97.68}  \\
\hline   
\end{tabular}}
\end{center}
\caption{Facial expression recognition result on Oulu-Casia and CK+ dataset. } 
\label{table:1}
\end{table}

\begin{table}[hb!]

\begin{center}{
\scalebox{0.8} {
\begin{tabular}{|l | c|c| }
\hline
 Method &Age on MORPH & Gender On Adience\\\hline  

Baseline &  84.2 & 87.6    \\
2MD-MTL  & 89.3&  90.8  \\
DA-2MD-MTL &  88.9 & 91.4 \\
Distill-Baseline& 84.9 & 87.9 \\
Distill-2MD-MTL& 89.8   & 91.2   \\
Distill-DA-2MD-MTL&  89.5 &  91.7 \\
DR- Distill-DA-2MD-MTL&  \textbf{90.3} &  \textbf{92.9} \\
\hline   
\end{tabular}}}
\end{center}

\caption{Age estimation performance on MORPHII and Gender Estimation on Adience.} 
\label{table:3}
\end{table}

\begin{figure}[h]

\centering

  \scalebox{0.5}{\includegraphics[scale=0.6]{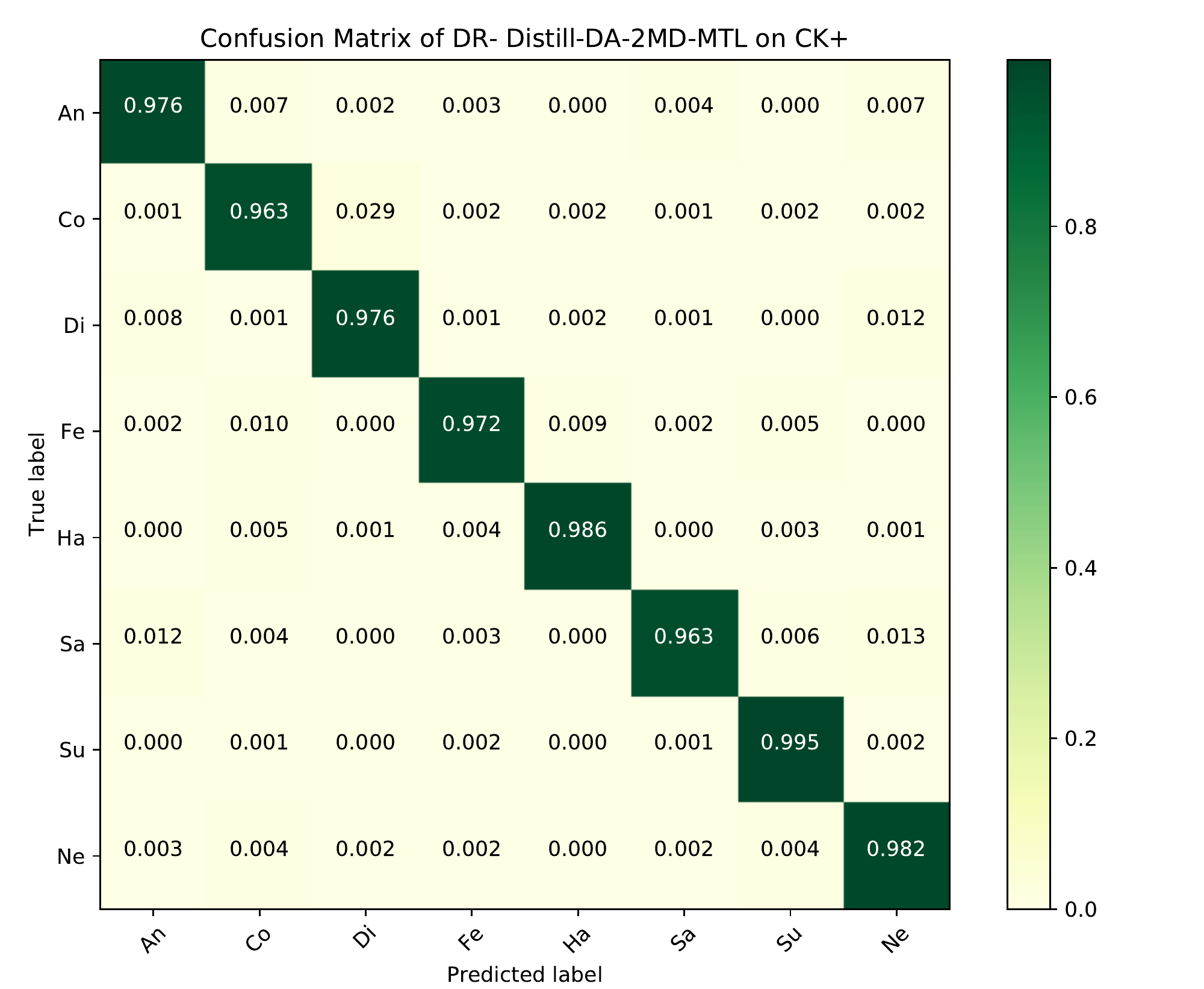}}
   \caption{Confusion Matrix from DR- Distill-DA-2MD-MTL on CK+., Casia(right). The darker the color, the higher the accuracy.}

\end{figure}

\begin{table}[h]

\begin{center}
\scalebox{0.75}{
      
\begin{tabular}{|l | c|c| c|}
\hline
 Method &Trained on CK+ & Trained on Casia & Trained on Both \\\hline  

Baseline & 33.1  & 35.2 & 39.8\\
2MD-MTL  & 35.6 & 38.3  & 47.0\\
DA-2MD-MTL   & 36.7 &38.5 & 47.3 \\
Distill-Baseline&  34.7 & 35.9 &  41.3 \\
Distill-2MD-MTL&  38.4 & 38.7 &  54.0 \\
Distill-DA-2MD-MTL& 38.1  & 38.5  &  54.2 \\
DR- Distill-DA-2MD-MTL& \textbf{38.9}  & \textbf{39.7} & \textbf{55.4} \\
\hline   
\end{tabular}}
\end{center}

\caption{Cross-Domain facial expression recognition result on FER 2013.   } 
\label{table:2}
\end{table}

\subsection{Cross Datasets Cross-Domain Evaluation}
\label{Semi-Supervised cross domain}
Not only our method benefits from all of the datasets to improve the results of the target dataset, but it is also capable of predicting the target labels on the domain of the auxiliary datasets. For validating these properties, we train our network same as Sec.~\ref{Semi-Supervised MTL}, except that we evaluate our network on FER2013. For training the network we use all CK+ dataset while training on CK+, and all Casia dataset while training on Casia also we use all Casia and CK+ dataset together as training set as the number of training image in each individual dataset were so low and the network could be easily get overfed. Results are provided in Table~\ref{table:2}, which shows that our method achieves significant results without seeing any labeled image of the target task in the domain of FER2013.

\section{conclusion}
We have proposed an end-to-end multi-dataset, multi-domain, and multi-task deep learning framework for joint facial expression, age, and gender estimation. The proposed scheme is able to exploit multiple datasets which the labels for different domains or tasks in the manner of semi-supervised learning. Hence, unlike the supervised multi-task network that needs expensive multiple labeled datasets, the proposed method is more efficiently trained. Using domain adaptation and data distillation, we were able to enhance the network generalization and solve the cross-domain adaptivity problem.


{\small
\bibliographystyle{ieee_fullname}
\bibliography{egbib}
}

\end{document}